\documentclass[journal]{IEEEtran}
\ifCLASSINFOpdf
\else
\fi

\usepackage{amsmath}
\usepackage{graphicx}
\usepackage{subfigure}
\usepackage{multirow}
\usepackage[colorlinks,
linkcolor=blue,
anchorcolor=blue,
citecolor=blue
]{hyperref}

\begin{document}
\title{DSMNet: Deep High-precision 3D Surface Modeling from Sparse Point Cloud Frames}


\author{Changjie Qiu, Zhiyong Wang, Xiuhong Lin, Yu Zang,~\IEEEmembership{Member,~IEEE}, Cheng Wang,~\IEEEmembership{Senior Member,~IEEE}, Weiquan Liu,~\IEEEmembership{Member,~IEEE}

\thanks{Manuscript received March, 2023. This work is supported by the China Postdoctoral Science Foundation (No.2021M690094); the FuXiaQuan National Independent Innovation Demonstration Zone Collaborative Innovation Platform (No.3502ZCQXT2021003). (\textit{Corresponding author: Cheng Wang and Weiquan Liu}.) }

    \thanks{C. Qiu, Z. Wang, X. Lin, Y. Zang, C. Wang and W. Liu,are with the Fujian Key Laboratory of Sensing and Computing for Smart Cities and Key Laboratory of Multimedia Trusted Perception and Efficient Computing, Ministry of Education of China, School of Informatics, Xiamen University, Xiamen 361005, China (e-mail: Qiuchangjie@stu.xmu.edu.cn; wangzy@stu.xmu.edu.cn; xhlinxm@stu.xmu.edu.cn; zangyu7@126.com; cwang@xmu.edu.cn; wqliu@xmu.edu.cn).}% <-this % stops a space

}
        % <-this % stops a space}


\markboth{IEEE GEOSCIENCE AND REMOTE SENSING LETTERS}%
{Shell \MakeLowercase{\textit{et al.}}: Bare Demo of IEEEtran.cls for IEEE Journals}



\maketitle
\begin{abstract}
Existing point cloud modeling datasets primarily express the modeling precision by pose or trajectory precision rather than the point cloud modeling effect itself. Under this demand, we first independently construct a set of LiDAR system with an optical stage, and then we build a HPMB dataset based on the constructed LiDAR system, a High-Precision, Multi-Beam, real-world dataset. Second, we propose an modeling evaluation method based on HPMB for object-level modeling to overcome this limitation. In addition, the existing point cloud modeling methods tend to generate continuous skeletons of the global environment, hence lacking attention to the shape of complex objects. To tackle this challenge, we propose a novel learning-based joint framework, DSMNet, for high-precision 3D surface modeling from sparse point cloud frames. DSMNet comprises density-aware Point Cloud Registration (PCR) and geometry-aware Point Cloud Sampling (PCS) to effectively learn the implicit structure feature of sparse point clouds. Extensive experiments demonstrate that DSMNet outperforms the state-of-the-art methods in PCS and PCR on Multi-View Partial Point Cloud (MVP) database. Furthermore, the experiments on the open source KITTI and our proposed HPMB datasets show that DSMNet can be generalized as a post-processing of Simultaneous Localization And Mapping (SLAM), thereby improving modeling precision in environments with sparse point clouds.

\end{abstract}

\begin{IEEEkeywords}
3D surface modeling, modeling evaluation, point cloud registration, point cloud sampling, post-processing of SLAM.
\end{IEEEkeywords}
\IEEEpeerreviewmaketitle
\section{Introduction}

\IEEEPARstart{O}{ver} the past few decades, most widely used LiDARs have centimeter-level range errors and visible point spacing due to low-beam. How to utilize such sparse point clouds for 3D surface modeling has become essential in various fields, such as autonomous driving, robot navigation, industrial manufacturing, etc. Existing point cloud modeling datasets (KITTI\cite{geiger2012we}, OXFORD\cite{9196884}, etc.) do not provide a unified and complete method to evaluate the modeling effect, whereas primarily use pose precision for modeling evaluation. However, due to the inhomogeneous surface point clouds of sparse point cloud frames, pose-based methods\cite{mur2015orb} will cause inevitable errors compared with the accurate modeling precision.

In this paper, under the above demand, first, we propose HPMB dataset, a High-Precision, Multi-Beam real-world, LiDAR dataset. HPMB consists of numerous low-precision LiDAR-scanned sequences with high-precision position and modeling ground truth. Second, we propose an evaluation criterion focusing on modeling effect. In detail, we utilize a simple similarity method to measure the difference between the low-precision LiDAR modeling results and high-precision LiDAR capture. Following this criterion, we can evaluate scene-level and object-level modeling precision.

\begin{figure}[t]
    \vspace{-0.5em}
  \centering
  \includegraphics[width=\columnwidth]{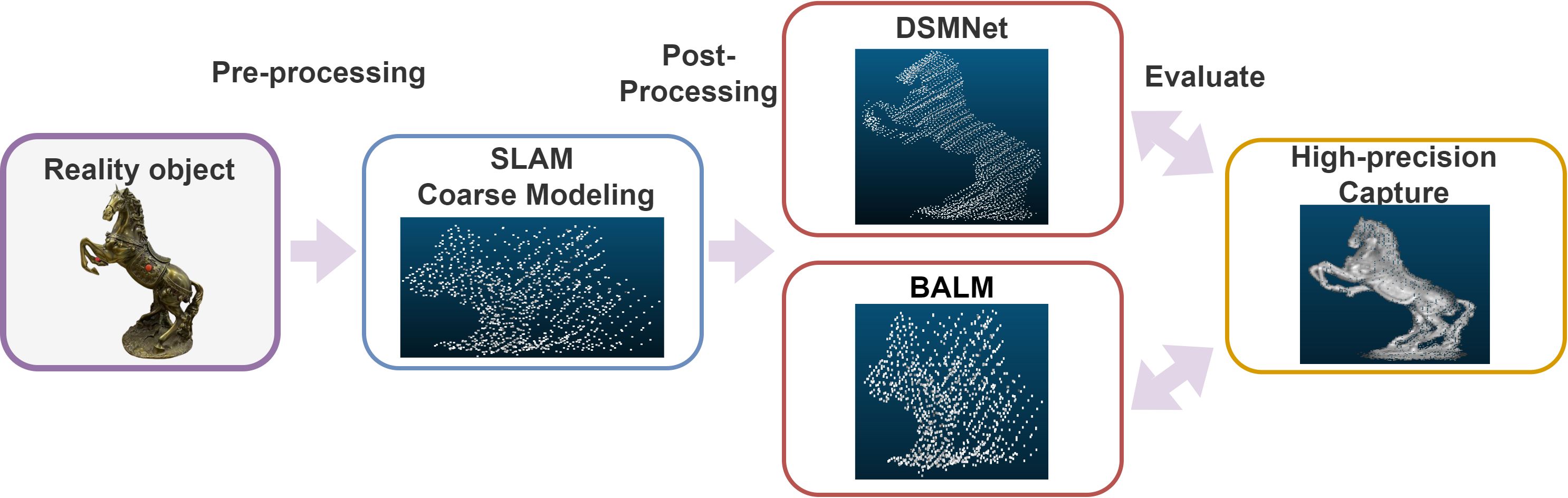}
   \caption{\textbf{Our proposed DSMNet performs accurate surface modeling of complex objects (horse in the HPMB dataset), better than BALM\cite{liu2021balm}. }}
   \label{f1}
   \vspace{-1.0em}
\end{figure}
Furthermore, %after our observation,
existing methods (BALM\cite{liu2021balm}, etc.) demonstrate ineffective in modeling complex object surfaces in sparse point clouds (see Fig. \ref{f1}). The problem of learning modeling of the environment is customarily known as Simultaneous Localization and Mapping (SLAM). Among these,  mapping tasks are often operated as post-processing, such as moving least squares, bundle adjustment and others\cite{triggs2000bundle}. Most methods rely on the matching features on the prior inter-frame registration results for modeling correction, which leads to low adaptability in sparse point clouds\cite{triggs2000bundle}. In this case, the shape characteristics of the object will become increasingly blurred and continue to disturb the inter-frame pose estimation.

Therefore, preliminary inter-frame pose estimation plays a vital role in this task, which is accomplished with point cloud registration (PCR). There are many effective traditional algorithms in the field of PCR. Besl et al. \cite{besl1992method} proposed a groundbreaking iterative optimization method based on point-wise distance. Dellenbach et al.\cite{dellenbach2022ct} utilizes point-based feature matching. Besides, methods based on deep learning demonstrated excellent results. Li et al.\cite{li2020iterative} proposed IDAM based on the idea of ICP, and Pan et al.\cite{pan2021robust} proposed GMCNet, which introduced the popular transformer in the matching framework.

However, the destination of the PCR is only to maximize the coincidence rate of both point clouds, but it do not handle the situation of significant noise and redundant surface points. In this scenario, it is desirable to filter out insignificant points, at which Point Cloud Sampling (PCS) is considered helpful. Equivalent to PCR, the most commonly used method is the traditional method farthest point sampling\cite{moenning2004intrinsic}, which preserves point clouds by iteratively picking the most faraway points. In the field of deep learning, S-Net\cite{dovrat2019learning} and SampleNet\cite{lang2020samplenet} both adopt the same training method that instructs a separate PCS module by training the downstream task.

In light of this, we propose a learning-based approach, DSMNet, while jointly optimizing the inter-frame pose estimation and the location of points. Instead of using simple surfel or plane methods, we use geometry-aware PCS to enhance the ability to obtain robust geometry features of sparse point clouds. As for feature matching, density-aware PCR achieves overlapping consistency for objects with complex structures. Additionally, we jointly optimize PCR and PCS to enhance feature fusion between modules. We utilize neighbor-focused PCR to transfer local density information to PCS by a point-wise weighting map, making PCS adaptable to uneven density. At the same time, we pass the geometric information as another weighting map of global-focused PCS to PCR, which handles the situation of low overlap and shape confusion. Overall, our main contributions are as follows:

• We independently construct a set of LiDAR system with an optical stage to contribute a high-precision, multi-beam, real-world dataset HPMB, containing over 3,000 points cloud frames with high-precision position and modeling information. Based on HPMB, we propose a unified and adaptable evaluation of the modeling precision method.

• We propose a novel learning-based framework, DSMNet, to jointly optimize geometry-aware PCS and density-aware PCR, which simultaneously learns implicit density and geometry information.

• Extensive experiments demonstrate that DSMNet outperforms previous state-of-the-art methods in PCS and PCR. Besides, DSMNet can be generalized as a post-processing of SLAM, which can precisely model complex objects in sparse point clouds.

\begin{figure}[t!] 
\vspace{-0.5em}
\centering
\includegraphics[width=\columnwidth]{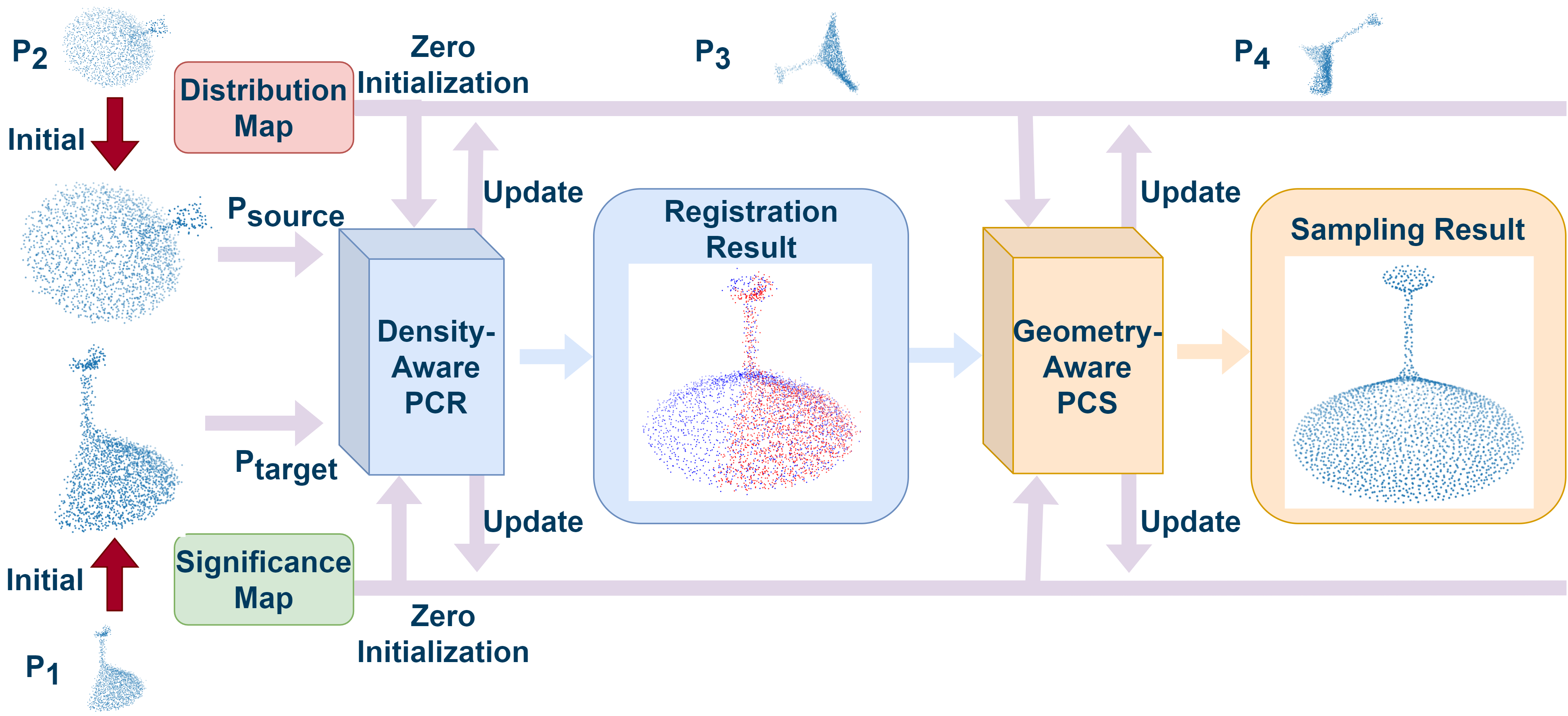}
\caption{\textbf{Overview of DSMNet:} DSMNet alternately processes point cloud frames through the Density-Aware PCR and the Geometry-Aware PCS module. In each alternation, DSMNet predicts the complete and precise point cloud obtained by the current processing, which continuously updates two point-wise weighting maps used for feature fusion.} %在overleaf上的名字
\label{fig:overall} 
\vspace{-1.0em}
\end{figure}%这个*可以实现跨双栏，切记前后都要加

\section{DSMNet}
Given a sequence of point cloud frames $I=\left\{P_{1}, \ldots, P_{m}\right\}$, where $P=\left\{p_{i} \in R^{3}\right\}_{i=1}^{n}$. 3D surface modeling from point cloud frames aims to obtain precise and complete modeling of corresponding objects in a unified world coordinate system. The architecture of DSMNet is shown in Fig. \ref{fig:overall}.

\subsection{Density-Aware Point Cloud Registration} \label{sec:PCR}
In the case of sparse point clouds, the locations of surface points are affected by LiDAR measurement errors. To address the noise generated by such errors, we follow the construction pattern of variational autoencoder (VAE) for feature extraction. This makes feature distribution close to the normal distribution, reducing the position error caused by normal noise. Moreover, we divide it into a global encoder and a point-wise neighborhood encoder. The details of the feature encoder follow the encoder design of Pointnet++\cite{qi2017pointnet++}.

\begin{figure}[t] 
\vspace{-0.5em}
\centering
\includegraphics[width=\columnwidth]{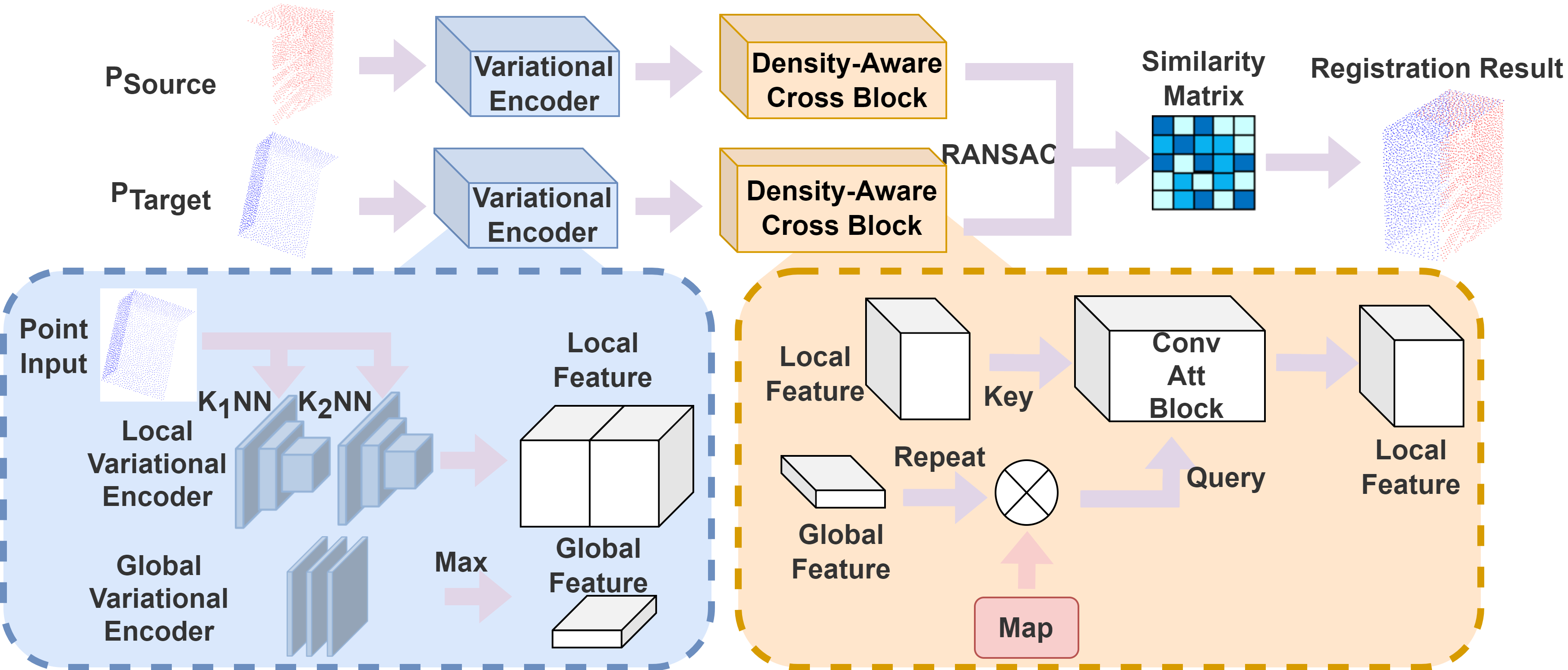}
\caption{\textbf{Overview of Density-Aware PCR module:} The module unifies the point-wise density feature through attention kernels of different scales of horizons. Given the source and target point clouds, the registered source point cloud is obtained through the similarity matrix calculated by RANSAC.} %在overleaf上的名字
\label{fig:registration} 
\vspace{-1.0em}
\end{figure}%这个*可以实现跨双栏，切记前后都要加

After obtaining global and point-wise local features, the PCR module adopts a novel local density consistent cross-attention module to combine global and local information (see Fig. \ref{fig:registration}). Specifically, we use local feature distribution as a query matrix and global feature distribution as a key matrix to query the distribution difference of neighborhoods in global information. Based on this, we use the proportion of the difference to mask the point-wise local features, which filter points with non-uniform densities. Additionally, to accommodate the densities of the corresponding point neighborhoods of the source and target point clouds, we aggregate local features using different scales of horizons. Finally, the relationship matrix $M$ between two point clouds is obtained by RANSAC\cite{chum2003locally}, and a point-wise significance map $R$ oriented to PCR, which is used for cross-module feature fusion, defined as follows:

\begin{small}
\begin{equation}
\vspace{-0.5em}
{R(p_{i})=\sum _{\mathbf{p_{i} \in p}} top5 (M_{i,1}, M_{i,2},\ldots,M_{i,n})}
\end{equation}
\end{small}

\noindent where $M$ is the similarity matrix between the source point cloud and the target point cloud. The significance map is the sum of five largest similar weights associated with per point on the similarity matrix, indicating the likelihood that the point belongs to an inline point of another point cloud during the registration process. The loss $L_{R}$ operated in the registration module is the distance between the predicted and ground truth registered point clouds, formulated as:
\begin{equation}
L_R=\frac{1}{N} \sum_i^N|(R_{gt} p_i+t_{gt})-(R_{pred} p_i+t_{pred})|
\vspace{-0.5em}
\end{equation}

\noindent where $R$ is the rotation matrix and $t$ is the translation vector of the source point cloud registered to the target point cloud.

\begin{figure}[t]
\vspace{-0.5em}
\centering
\includegraphics[width=\columnwidth]{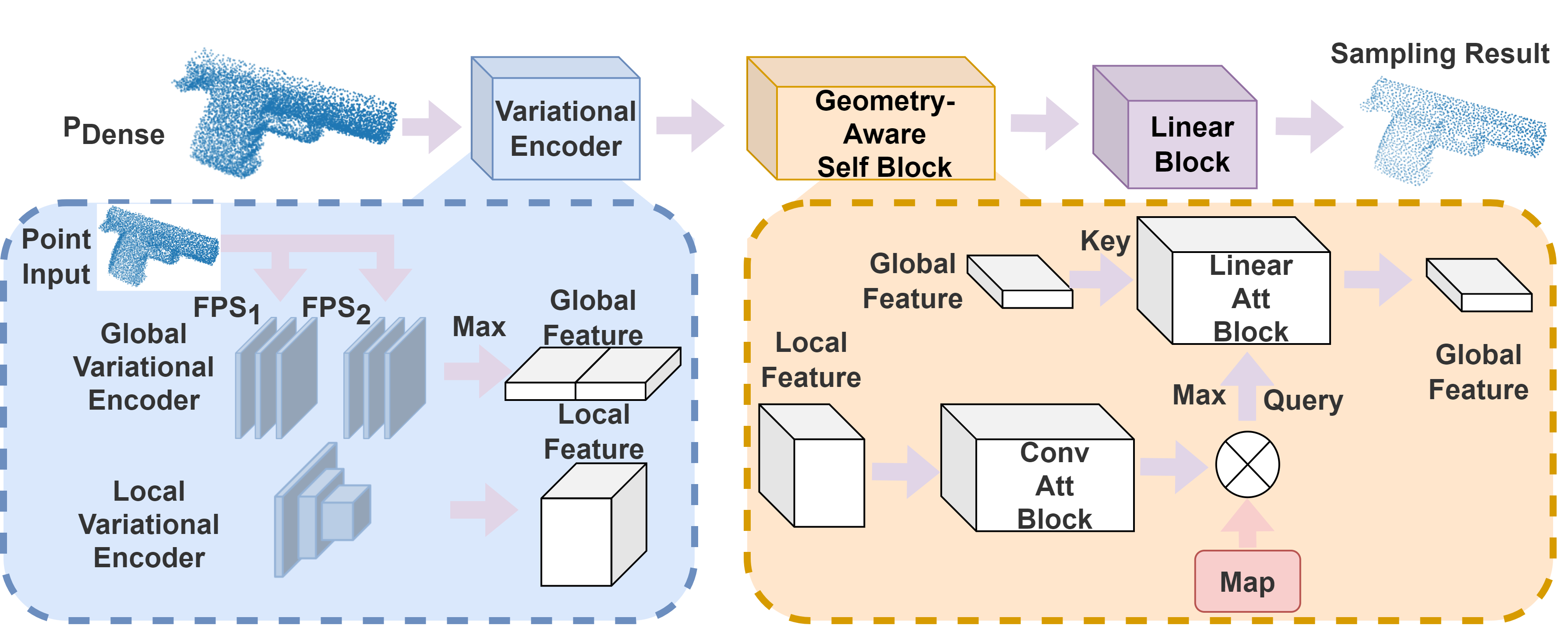}
\caption{\textbf{Overview of Geometry-Aware PCS module:} The module explores the implicit shape feature through attention kernels at different resolution scales. Given a point cloud, a complete, precise sampled result is obtained.} %在overleaf上的名字
\vspace{-1.0em}
\label{fig:sample} 
\end{figure}%这个*可以实现跨双栏，切记前后都要加

\subsection{Geometry-Aware Point Cloud Sampling} \label{sec:PCS}

Divergent from the PCR module, PCS utilizes a geometry consistent self-attention module for geometric feature enhancement. Afterward, an accurate and complete point cloud $O$ with uniform density is obtained by uncomplicated point-wise position regression. Each attention kernel further explores the implicit global shape information of point clouds by fusing point-wise features corresponding to point clouds with different scales of resolution. The details are shown in Fig. \ref{fig:sample}.

The significance map $S$ of PCS is obtained from the minimum distance calculation between sampled and original point clouds, which is defined as follows:
\begin{small}
\begin{equation}
{S(p_{i})=\min_{\mathbf{p_{i}} \in \mathcal{P},\mathbf{o_{i}} \in \mathcal{O}} \left\|p_{i}-o_{j}\right\|_{2}^{2}}
\end{equation}
\vspace{-0.5em}
\end{small}

The significance map $S$ is represented as the point-wise importance of the global shape, measured by point-wise minist distance. The loss $L_{S}$ operated in the sampling module, the traditional Chamfer Distance\cite{lang2020samplenet}, can be formulated as: 

\begin{small}
\begin{equation}
\begin{array}{l}
{L_{S}(P, O)=\frac{1}{n} \sum_{i}^{n} \min_{\mathbf{p \in P, o_{i} \in O}}\left\|\mathbf{o}_{i}-\mathbf{p}\right\|_{2}^{2}} \\
{+ \frac{1}{n} \sum_{i}^{n} \min _{\mathbf{o} \in O, p_{i} \in P}\left\|\mathbf{p}_{i}-\mathbf{o}\right\|_{2}^{2}}
\end{array}
\label{eq:chamfer}
\vspace{-0.5em}
\end{equation}
\end{small}

\subsection{Cyclic Optimization Based on Point-wise Weighting Map}

Our DSMNet is trained end-to-end and optimizes the PCR and PCS modules alternately (see Fig. \ref{fig:overall}). Initially, we utilize the PCR module to register the point clouds to a unified world coordinate system and the PCS module to sample the registered point clouds for a concise result. In the next iteration, we utilize the sampled point cloud of the previous iteration as the target point cloud for registration.

However, cyclic modeling will lead to the accumulation of errors. In order to effectively suppress this situation, we utilize two point-wise weighting maps for reducing the attention weights of low utility points generated by previous iterations. Precisely, we not only utilize the significance map introduced in Sec. \ref{sec:PCR} and \ref{sec:PCS}, but also construct a point-wise neighborhood distribution map $D$. After computing global and point-wise local feature distributions, $M$ points are randomly sampled from each local feature distribution. Based on these points, the probability belonging to the global feature distribution is calculated as the value of the point-wise distribution map. The specific calculation method is as follows:

\begin{small}
\begin{equation}
\vspace{-0.5em}
\begin{split}
\begin{aligned}
{D(p_{i})=\sum_{a_j \in dl(p_{i})}^{M}({-\frac{(a_j-\mu)^{2}}{2 \sigma^{2}}-\log (\sigma)-\log (\sqrt{2 \pi})})}
\end{aligned}
\end{split}
\end{equation}
\end{small}

\noindent where $\mu$ and $\sigma$ is the mean and variance of the global feature distribution. $dl(\cdot)$ is the samping point set of the local feature distribution. M is the number of sampling points.

The point-wise distribution map, containing features about the neighborhood density, assigns task-irrelevant information for module optimization. The significance map provides task-oriented information and the geometry importance of the local shape in the global shape. Through the cyclic, the map obtained by PCR assists PCS in performing a preliminary sampling based on density. PCS assists PCR in screening geometry outliers. It is worth noting that since there are no corresponding results to calculate in the first iteration, we replace the weighting map with two matrices of all ones.

\begin{figure}[t!]
\vspace{-0.5em}
  \centering
  \includegraphics[width=\columnwidth]{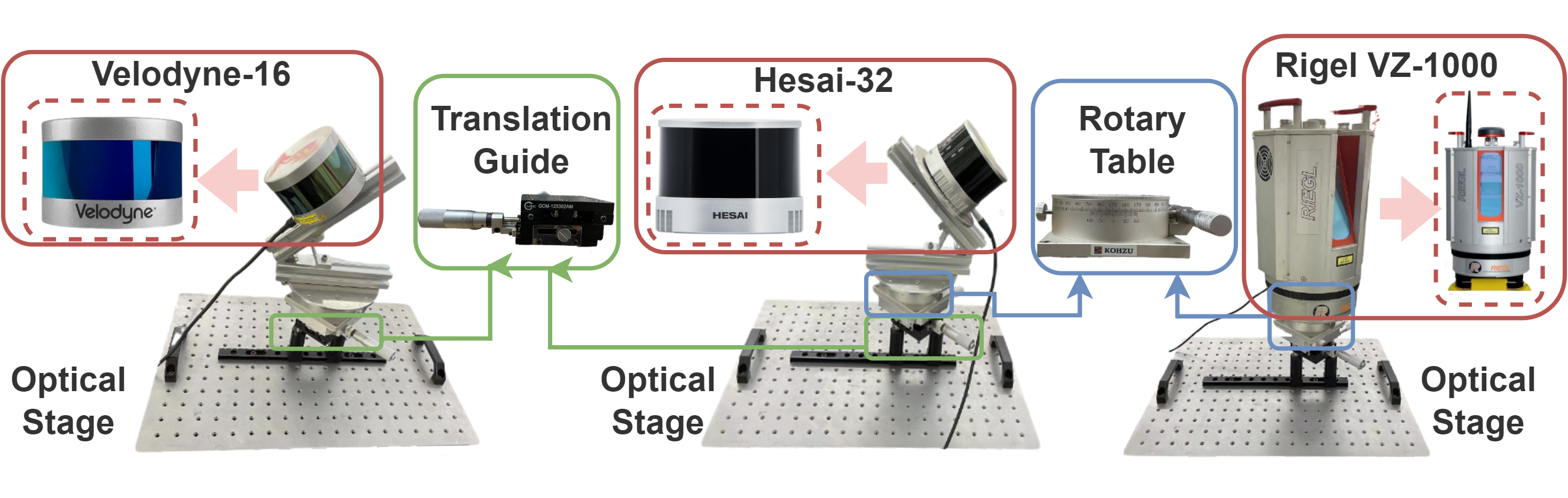}
  
   \caption{\textbf{Overview of HPMB LiDAR System:} The system consists of an optical stage, a milliradian-level precision translation guide, a millimeter-level precision rotary table, and different LiDARs.}
   \label{fig:HPMB}
\vspace{-0.5em}
\end{figure}

\section{HPMB Dataset}
\label{sec:HPMB}
%Towards an effort to address the challenge of high-precision 3D surface modeling of complex objects, we contribute HPMB, a high-precision, multi-beam, real-world dataset, to the community. Within HPMB, three completely static objects are captured by two types of LiDARs with different precision and beams. Specifically, we build an entirely bound LiDAR system consisting of an optical stage, a high-precision translation guide, and a rotary table. (see Fig. \ref{fig:HPMB}). 

%Towards an effort to address the challenge of high-precision 3D surface modeling of complex objects from sparse point clouds, we contribute HPMB dataset, a high-precision, multi-beam, real-world dataset, to the community.
We independently construct a set of LiDAR system with an optical stage, containing a millimeter-level precision translation guide and a milliradian-level precision rotary table to collect data, which is illustrated in Fig. \ref{fig:HPMB}.

• \textbf{Data Acquisition.}  During the capture process, multiple complex structured objects were placed in a static scene at two meters from the LiDAR system. We operate the high-precision translation guide and rotary table to move the low-precision LiDARs to simulate realistic multi-view modeling. After each movement, we record the ground truth of high-precision rotation and translation at the current time node. At the same time, under a unified world coordinate system, we use a millimeter-level precision LiDAR to capture, regarded as the ground truth of high-precision global modeling.

In summary, HPMB provides 3,600 time node, three kinds of LiDARs (Velodyne-16, Hesai-32 and Rigel VZ-1000), and 10 kinds of complex objects (Horse, Cabinet, Sofa, etc.).

% \usepackage{graphicx}
\begin{table}[t]\LARGE\renewcommand\arraystretch{1.5}
\vspace{-0.5em}
\centering
\caption{Comparing HPMB with existing datasets.}
\label{tab:precision}
\resizebox{\columnwidth}{!}{%
\begin{tabular}{c|c|c|c}
\hline
Dataset    & Translation Precision & Rotation Precision & Modeling Precision \\ \hline
OXFORD\cite{9196884}     &$\geq 1cm$                  & 0.02°              & 20mm               \\ \hline
KITTI\cite{geiger2012we}      & $\geq 1cm$                   & 0.15°              & 20mm               \\ \hline
3DMatch\cite{zeng20173dmatch}    & 2cm                   & 0.1°               & ×               \\ \hline
\textbf{HPMB(Ours)} & \textbf{0.01cm}                & \textbf{0.01°}              & \textbf{5mm}           \\ \hline
\end{tabular}%
}
\vspace{-0.8em}
\end{table}

% \begin{table}[ht]\LARGE\renewcommand\arraystretch{1.5}
% \begin{center}
% \vspace{-0.5em}
% \caption{\textbf{Comparing HPMB with existing datasets.}}
% \label{tab:precision}
% \resizebox{\linewidth}{!}{
% \begin{tabular}{c c c C} 
%  \hline
%  Dataset & Translation Precision & Rotation Precision & Modeling precision\\
%  \hline
% Oxford\cite{9196884} & \geq 1cm & \geq 0.02 ° & 20mm \\
% KITTI\cite{geiger2012we} & \geq 1cm & \geq 0.15 ° & 20mm\\
% 3DMatch\cite{zeng20173dmatch} & 2cm & 0.1° & × \\
% HPMB (Ours) & \textbf{0.01cm} & \textbf{0.01°} & \textbf{5mm}\\
%  \hline
% \end{tabular}
% }
% \end{center}
% \vspace{-0.5em}
% \end{table}

• \textbf{Data Characteristic.} 
Table \ref{tab:precision} presents statistics of our HPMB dataset in comparison to other publicly available datasets. Our HPMB dataset has the following advantages: (1) HPMB consists of over 3,000 high-precision optical flow ground truth, while other registration-oriented or SLAM-oriented datasets usually have limited precision. (2) HPMB is comprised of multiple LiDARs with varying beams and precision. Most current datasets lack such high modeling precision. (3) On the basis above two, we can calculate the similarity between the modeling results of low-precision LiDAR and the capture of high-precision LiDAR at any time node, which can be regarded as a unified modeling effectiveness evaluation method. Existing datasets use pose precision to replace the evaluation of modeling precision, which has low precision and inevitable error when the point cloud is sparse.

%The capture process of each object is divided into 18 groups. The translation guide is moved 5cm between each group, the rotary table is rotated five times by 5° within each group, and each state is recorded as a particular time node. At the same time, we capture the same object at the exact location with a millimeter-precision level Rigel scanner. In this case, we have a sequential time node of low-precision LiDAR, where each has high-precision rotation, translation, and modeling ground truth. On this basis, we can intuitively evaluate the modeling results at each time node with the high-precision capture, which can be generalized as a unified and adaptable method for evaluating modeling effects. In general, our dataset has many advantages over other datasets.

%• \textbf{high-precision}. Due to the high precision of the system equipment and the Rigel laser scanner, each time point of HPMB has high-precision rotation, translation, and modeling ground truth. The previous real-world datasets(kitti\cite{geiger2013vision,geiger2012we}, etc.) lack such precision. (see Table. \ref{tab:dataset precision}).

%• \textbf{Multi-beam}. To address the challenge of modeling by sparse point clouds, we operate two low-precision, low-beams LiDARs (Velodyne-16 and Hesai-32) to capture the sparse point cloud of objects. Furthermore,  we use a high-beams LiDAR scanner (Rigel VZ-1000) for scanning to obtain high-precision modeling ground truth.

\begin{figure}[t]
  \vspace{-0.5em}
  \centering
  \includegraphics[width=\columnwidth]{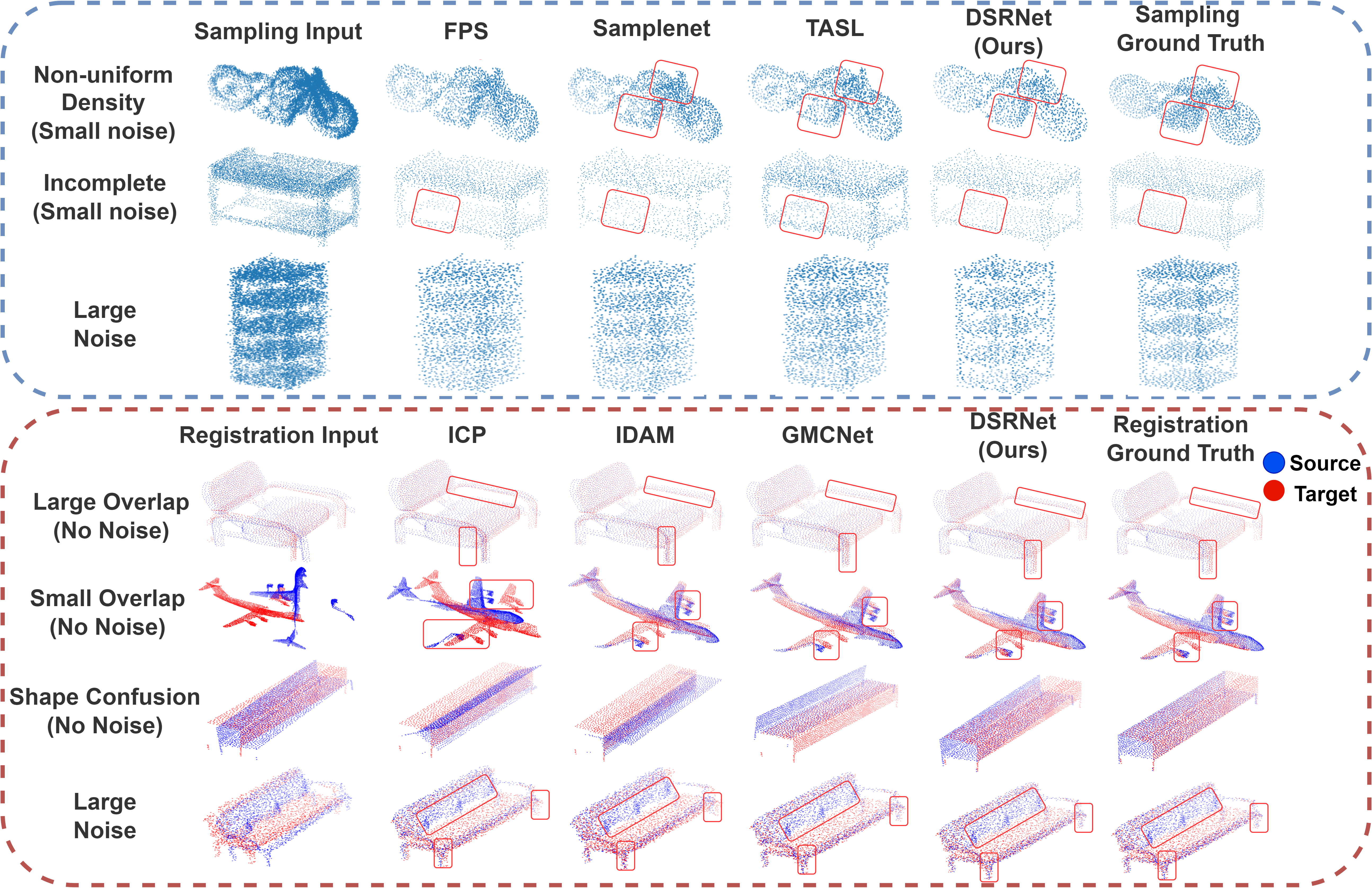}
  \caption{\textbf{Qualitative results on the MVP-SP and MVP-RG datasets.} The upper part is PCS and the lower part is PCR, which proves DSMNet can handle various challenging conditions (Large Noise, Moving, etc.).}
  \label{fig:MVP}
  \vspace{-0.5em}
\end{figure}

\section{Experiments}
In line with previous methods, like VRCNet \cite{pan2021variational} and GMCNet \cite{pan2021robust}, we evaluate PCR by computing isotropic rotation errors and translation errors: $E_{R} = \arccos \left(\frac{1}{2}\left(\operatorname{tr}\left(R_{G_T}^{-1} \cdot R_{\text {pred }}\right)-1\right)\right)$, $E_{t} =|| t_{G T}-\left.t_{\text {pred }}\right||_2$. Besides, we evaluate PCS by computing the Chamfer Distance (Eq. (\ref{eq:chamfer})), which is also utilized as the evaluation method of modeling precision in the HPMB dataset.
\subsection{Point Cloud Registration and Sampling}
\label{subsection:MVP dataset}
MVP \cite{pan2021variational} is a multi-view partial point cloud dataset, consisting of over 100,000 complete and incomplete point clouds in 16 categories, which renders partial 3D shapes from 26 uniformly distributed camera poses for 3D CAD models.

% \usepackage{graphicx}
\begin{table}[t]\small\renewcommand\arraystretch{1}
\vspace{-0.5em}
\centering
\caption{Quantitative comparison of PCR in MVP-RG dataset.}
\label{tab:registration}
\resizebox{\columnwidth}{!}{%
\begin{tabular}{c|c|c}
\hline
Method  & Rotation Error $(E_{R})$      & Translation Error $(E_{t})$     \\ \hline
DeepGMR\cite{yuan2020deepgmr} & 43.74° & 0.353 \\ \hline
GMCNet\cite{pan2021robust}  & 16.57° & 0.174 \\ \hline
IDAM\cite{li2020iterative}    & 24.35° & 0.280 \\ \hline
RPMNet\cite{yew2020rpm}  & 22.20° & 0.174 \\ \hline
\textbf{DSMNet(Ours)}  & \textbf{14.17°} & \textbf{0.158} \\ \hline
\end{tabular}%
}
\vspace{-0.8em}
\end{table}

%As for point cloud registration, we follow the construction method of the MVP dataset\cite{pan2021variational}, we built a challenging multi-view registration dataset (MVP-RG) for PCR. In the MVP-RG dataset, each point cloud contains 2,048 points, Based on an index of sufficient overlapping regions being detected, pairs of point clouds were selected for the same object. In total, our MVP-RG dataset consists of 7,600 partial point cloud pairs from sixteen categories, divided into a training set (6,400 samples) and a test set (1,200 samples). We compared our method with the following deep learning methods: IDAM\cite{li2020iterative}, GMCNet\cite{pan2021robust}, DeepGMR\cite{yuan2020deepgmr}, etc. (see Table. \ref{tab:registration}). Experiments prove that our algorithm has a favorable effect on the MVP-RG dataset, which means our registration module exhibits excellent performance on challenging, low-overlap partial point clouds.

Following the construction method of the MVP dataset\cite{pan2021variational}, we construct a challenging multi-view registration dataset (MVP-RG) for PCR. In summary, MVP-RG dataset consists of 7,600 partial point cloud frame pairs from sixteen categories. We compared our DSMNet with existing superior methods (see Table. \ref{tab:registration}). Experiments demonstrate that our DSMNet has a favorable effect on the MVP-RG dataset, which represents our PCR module exhibiting excellent performance on challenging, low-overlap partial point clouds (see Fig. \ref{fig:MVP}).

\begin{figure}[t]
\vspace{-0.5em}
  \centering
  \includegraphics[width=\columnwidth, height=5cm]{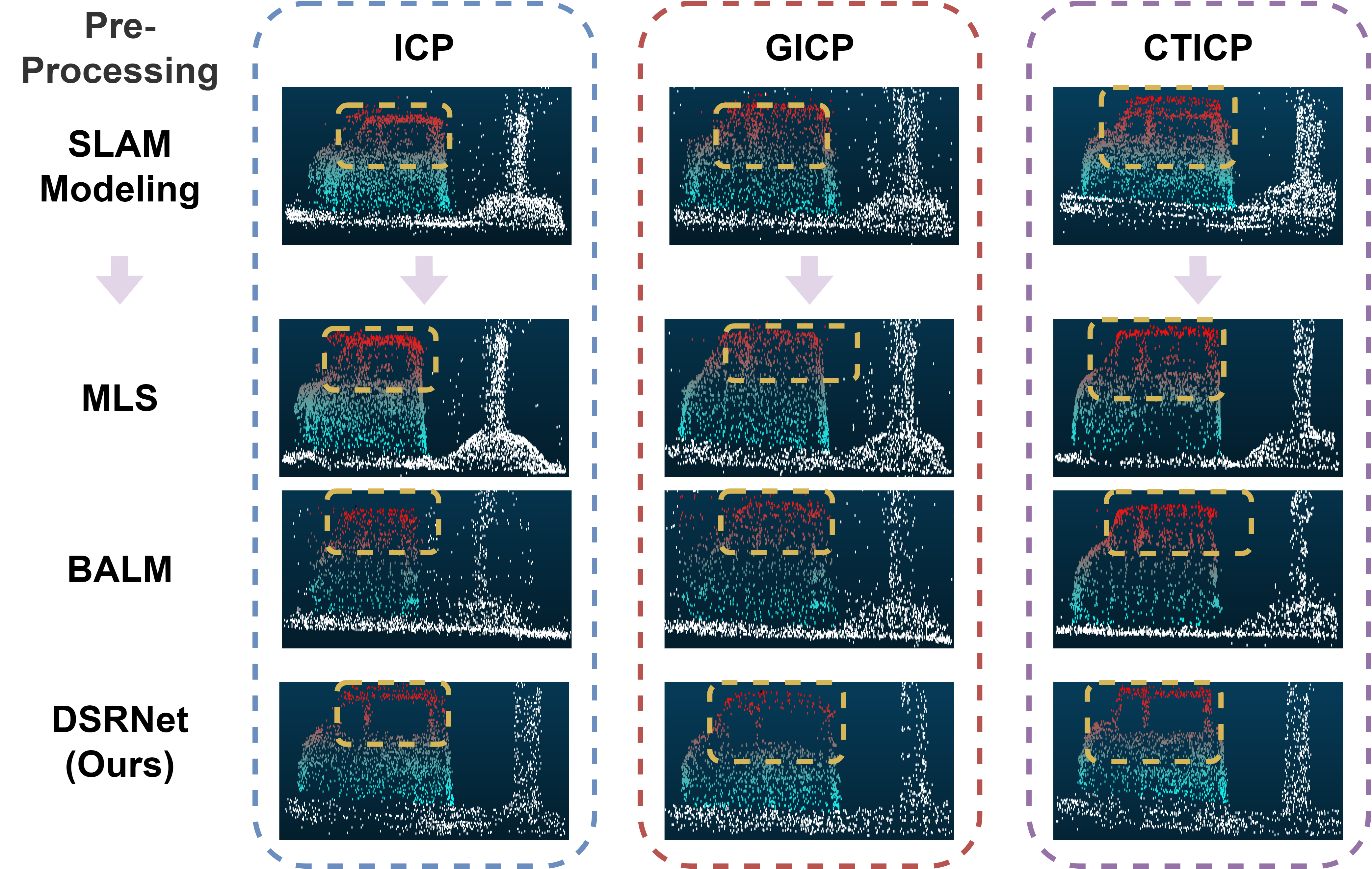}
   \caption{
\textbf{Qualitative modeling results on the KITTI dataset.} DSMNet can be used as a SLAM post-processing algorithm to achieve high-precision surface modeling of complex objects (Cars, Pillars, etc.)}
   \label{fig:qualitative-kitti}
   \vspace{-0.5em}
\end{figure}

\begin{table}[t]\small\renewcommand\arraystretch{1}
\centering
\vspace{-0.5em}
\caption{Quantitative comparison of PCS in MVP-SP dataset.}
\label{tab:sample}
\resizebox{\columnwidth}{!}{%
\begin{tabular}{c|c}
\hline
\multirow{2}{*}{Method}          & Modeling Precision  \\  & (Chamfer Distance F-Score@1$\%$) \\ \hline
FPS\cite{moenning2004intrinsic}             & 0.507                               \\ \hline
SampleNet\cite{lang2020samplenet}       & 0.5921                              \\ \hline
Local SampleNet & 0.5421                              \\ \hline
TASL\cite{lin2022task}            & 0.6410                              \\ \hline
S-Net\cite{dovrat2019learning}           & 0.3954                              \\ \hline
\textbf{DSMNet(Ours)}    & \textbf{0.6610}                              \\ \hline
\end{tabular}%
}
\vspace{-0.8em}
\end{table}

% \begin{table}[ht]\small\renewcommand\arraystretch{1}
% \begin{center}
% \vspace{-1.0em}
% % \resizebox{\linewidth}{!}{
% \caption{\textbf{Quantitative comparison (F-Score@1$\%$) of PCS in MVP-SP dataset.}}
% \label{tab:sample}
% \begin{tabular}{ c c} 
%  \hline
%  Method & \qquad \qquad Chamfer F1 \qquad \qquad \qquad   \\
%  \hline
% FPS & 0.507 \\
% SampleNet\cite{lang2020samplenet} & 0.5921 \\
% Local SampleNet & 0.5421 \\
% TASL\cite{lin2022task} & 0.6410  \\
% S-Net\cite{dovrat2019learning} & 0.3954\\
% DSMNet (Ours) & \textbf{0.6710} \\
%  \hline
% \end{tabular}
% % }
% \end{center}
% \vspace{-0.5em}
% \end{table}

%Due to the specificity of the task, we did not perform point cloud completion as used in the original MVP dataset. Given a partial point cloud, select other n angles point clouds belonging to the same object to stack and regard the complete point cloud of the object as the ground truth. In summary, our MVP-SP dataset consists of 104,000 point clouds in 16 categories, split into a training set (62,400 samples) and a test set (41,600 samples). We compare our method with traditional sampling algorithms such as FPS\cite{moenning2004intrinsic}, as well as learning-based deep learning algorithms SampleNet\cite{lang2020samplenet}, TASL\cite{lin2022task}, etc. (see Table \ref{tab:sample}). Experimental results demonstrate that our framework samples the MVP-SP dataset best, showing the superiority of our sampling module in the case of non-uniform densities and incomplete point clouds.

Unlike the point cloud completion (PCC) introduced in the original MVP dataset, we construct a challenging sampling dataset MVP-SP based on the data of the PCC. Specifically, for each object we randomly select several frames from the corresponding set of multi-view partial point cloud frames. On this basis, we stack all the selected frames, which is regarded as the input of PCS. Subsequently, the complete and concise point cloud provided by the PCC is used as the ground truth. In summary, our MVP-SP dataset consists of 104,000 point cloud frames of 16 categories. Experiments demonstrates that our DSMNet outperforms SOTA methods in the MVP-SP dataset, showing the superiority of our PCS module in the case of non-uniform densities and incomplete point clouds (see Fig. \ref{fig:MVP} and Table \ref{tab:sample}).

\subsection{Post-processing of Simultaneous Localization and Mapping}

The KITTI odometry dataset\cite{geiger2012we} consists of 22 independent sequences captured during driving over various road environments. Sequences 00-10 (23,201 scans) are provided with ground truth poses obtained from the IMU/GPS readings.

% Because we focus more on the effect of modeling than comparing the precision of trajectory coincidence, we use DSMNet as a post-processing of the SLAM process. Specifically, We compare our approach based on several classic LiDAR odometry estimation methods in slam, ICP\cite{besl1992method}, G-ICP\cite{koide2021voxelized}, CT-ICP\cite{dellenbach2022ct}. On this basis, we tested several post-processing algorithms, MLS\cite{liu2021deep}, BA\cite{liu2021balm}. For a more intuitive comparison effect, we selected some objects (Cars, Grass) from the KITTI dataset for visual comparison. As can be seen from the visualization, our framework also obtains excellent modeling precision on complex structures.

Rather than comparing the precision of trajectory in the KITTI dataset, we focus more on the effect of modeling. Specifically, we use several classic LiDAR odometry methods of SLAM as preprocessing. Based on these prior poses, we tested our DSMNet and several post-processing algorithms for fine modeling. To achieve an intuitive comparison, we selected a special and complex object, car, from the KITTI dataset for visual comparison (see Fig. \ref{fig:qualitative-kitti}). As the visualization shows, our framework also obtains excellent modeling precision on complex structural objects in the case of high noise or poor prior results. While in the environment of good prior and high signal-to-noise ratio point clouds, our algorithm achieves a similar effect comparable to these algorithms. 
% \begin{figure}[ht] 
% \centering
% \includegraphics[width=\columnwidth]{HPMM res.png}
% \caption{Visual comparison of multi-beam LiDAR modeling effects in HPMB dataset. The process setup is the same as the test on the KITTI dataset} %在overleaf上的名字
% \label{fig:sequential} 
% \end{figure}%这个*可以实现跨双栏，切记前后都要加

\begin{figure}[t]
\vspace{-0.5em}
\centering
\includegraphics[width=\columnwidth]{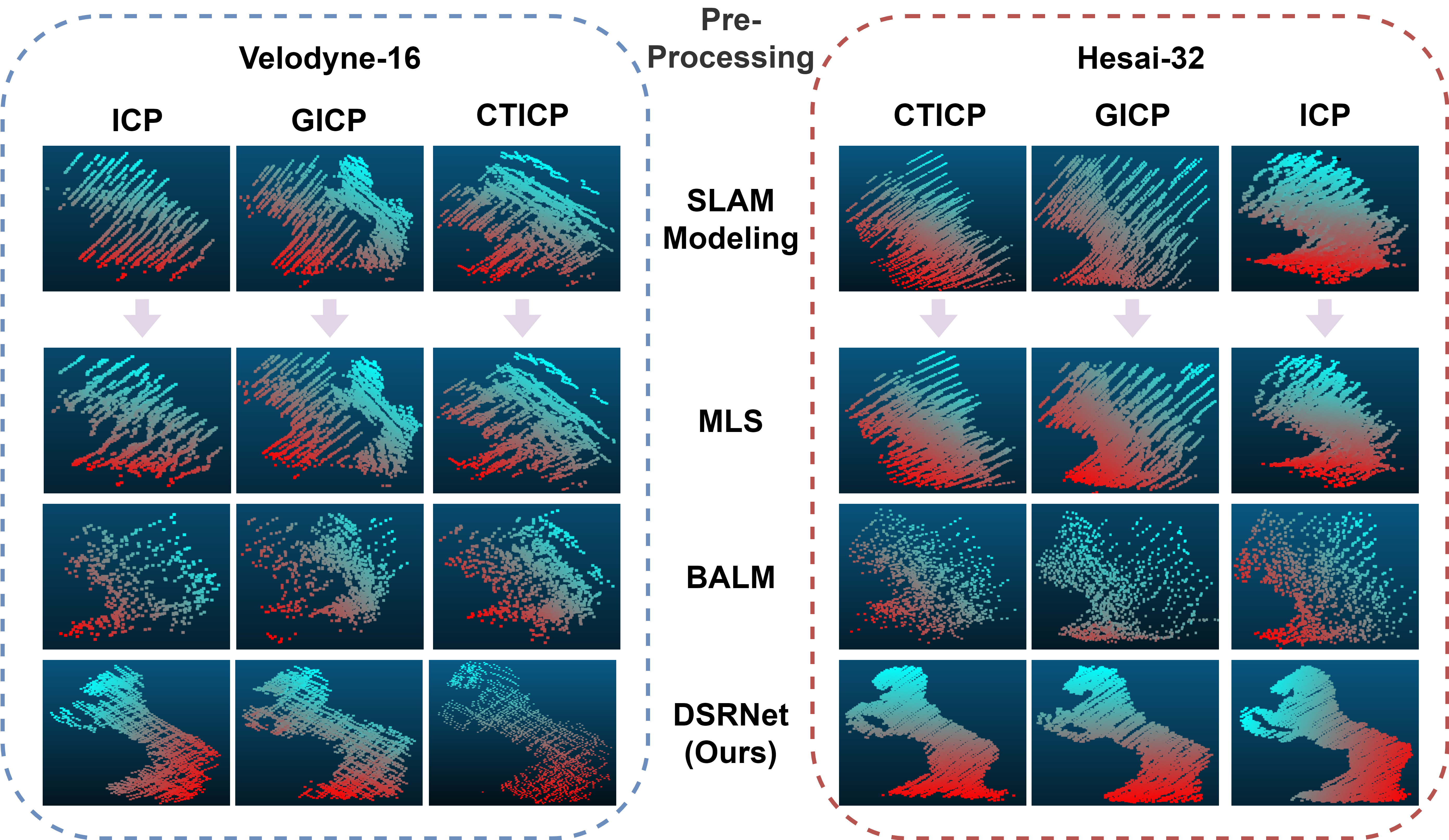}
\caption{\textbf{Qualitative modeling results on the HPMB dataset by different methods.} DSMNet can guarantee the integrity of the shape of the complex object (horse) in the case of sparse point clouds.} %在overleaf上的名字
\label{fig:sequential} 
\vspace{-1.0em}
\end{figure}%这个*可以实现跨双栏，切记前后都要加
However, due to the lack of methods to evaluate the effect of modeling in the KITTI dataset, we can only achieve qualitative experiments. As introduced in Sec. \ref{sec:HPMB}, our HPMB dataset utilizes a unified modeling effect evaluation method to realize quantitative experiments at the object level. Additionally, we also achieve qualitative experiments like the KITTI dataset.

The visual results shed light on the more lower precision of LiDAR, the more worse modeling quality due to sparsity and cumulative errors (see Fig. \ref{fig:sequential}). In this particular scenario, experiments demonstrate that our DSMNet effectively alleviates this situation and guarantees highly precise point cloud modeling results for complex object. For a more intuitive comparison effect, quantitative experiments demonstrate that DSMNet outperforms SOTA methods of modeling precision in our HPMB dataset (see Table \ref{tab:HPMB}).

% \usepackage{multirow}
% \usepackage{graphicx}

% \begin{table}[ht]\small\renewcommand\arraystretch{1}
% \vspace{-0.5em}
% \begin{center}
% \caption{\textbf{Quantitative comparison of modeling effect of Velodyne-16 LiDAR for complex objects on HPMB dataset.}}
% \resizebox{\linewidth}{!}{
% \begin{tabular}{ c c C} 
%  \hline
%  PreProcessing Method & PostProcessing Method & Modeling Precision ((F-Score@1$\%$) \\
%  \hline
%  & MLS \cite{liu2021deep} & 0.017 \\
%  ICP\cite{besl1992method} & BALM\cite{liu2021balm} & 0.024 \\
%  & DSMNet (Ours) & 0.321 \\
%   \hline
%  & MLS & 0.103 \\
%  G-ICP & BALM & 0.142 \\
%  & DSMNet (Ours) & 0.341 \\
%   \hline
%  & MLS & 0.079 \\
%  CT-ICP & BALM & 0.046 \\
%  & DSMNet (Ours) & 0.411 \\
%  \hline
% \end{tabular}
% }
% \label{tab:HPMB}
% \end{center}
% \vspace{-1em}
% \end{table}

%It is obvious from the visualization results that the lower precision of LiDAR, the worse quality of the modeled point clouds due to large LiDAR measurement errors and larger registration and sampling errors. However, experiments show that DSMNet effectively alleviates this situation and guarantees highly accurate point cloud modeling results for complex object geometries in the presence of sparse point clouds.

\section{Conclusion}

In this letter, we independently construct a set of LiDAR system with an optical stage to contribute a high-precision, multi-beam, real-world dataset HPMB. Based on HPMB, we propose a method to evaluate modeling precision, which fills the gap that the existing datasets cannot evaluate modeling quality. Besides, we propose a novel learning-based joint framework, DSMNet. DSMNet jointly optimizes density-aware PCR and geometry-aware PCS for modeling complex objects of sparse point cloud. Experiments show that DSMNet achieves the best results in PCS and PCR, and can be generalized as SLAM post-processing to solve the difficult challenge of high-precision modeling in complex environments.

\begin{table}[t]\small\renewcommand\arraystretch{1}
\centering
\vspace{-0.5em}
\caption{Quantitative comparison of modeling effect of Velodyne-16 LiDAR on HPMB dataset.}
\label{tab:HPMB}
\resizebox{\columnwidth}{!}{%
\begin{tabular}{c|c|c}
\hline
\multirow{2}{*}{PreProcessing}          & \multirow{2}{*}{Method}       & Modeling Precision \\ 
 & & (Chamfer Distance F-Score@1$\%$)\\ \hline
\multirow{3}{*}{ICP\cite{besl1992method}}   & MLS\cite{liu2021deep}          & 0.017                             \\ \cline{2-3} 
                       & BALM\cite{liu2021balm}         & 0.024                             \\ \cline{2-3} 
                       & \textbf{DSMNet(Ours)} & \textbf{0.321}                             \\ \hline
\multirow{3}{*}{GICP\cite{segal2009generalized}}  & MLS          & 0.103                             \\ \cline{2-3} 
                       & BALM         & 0.142                             \\ \cline{2-3} 
                       & \textbf{DSMNet(Ours)} & \textbf{0.341}                             \\ \hline
\multirow{3}{*}{CTICP\cite{dellenbach2022ct}} & MLS          & 0.079                             \\ \cline{2-3} 
                       & BALM         & 0.046                             \\ \cline{2-3} 
                       & \textbf{DSMNet(Ours)} & \textbf{0.411}                             \\ \hline
\end{tabular}%
}
\vspace{-1.2em}
\end{table}

\bibliographystyle{IEEEtran}
\bibliography{mybibfile}

\begin{thebibliography}{10}
\providecommand{\url}[1]{#1}
\csname url@samestyle\endcsname
\providecommand{\newblock}{\relax}
\providecommand{\bibinfo}[2]{#2}
\providecommand{\BIBentrySTDinterwordspacing}{\spaceskip=0pt\relax}
\providecommand{\BIBentryALTinterwordstretchfactor}{4}
\providecommand{\BIBentryALTinterwordspacing}{\spaceskip=\fontdimen2\font plus
\BIBentryALTinterwordstretchfactor\fontdimen3\font minus
  \fontdimen4\font\relax}
\providecommand{\BIBforeignlanguage}[2]{{%
\expandafter\ifx\csname l@#1\endcsname\relax
\typeout{** WARNING: IEEEtran.bst: No hyphenation pattern has been}%
\typeout{** loaded for the language `#1'. Using the pattern for}%
\typeout{** the default language instead.}%
\else
\language=\csname l@#1\endcsname
\fi
#2}}
\providecommand{\BIBdecl}{\relax}
\BIBdecl

\bibitem{geiger2012we}
A.~Geiger, P.~Lenz, and R.~Urtasun, ``Are we ready for autonomous driving? the
  kitti vision benchmark suite,'' in \emph{2012 IEEE conference on computer
  vision and pattern recognition}.\hskip 1em plus 0.5em minus 0.4em\relax IEEE,
  2012, pp. 3354--3361.

\bibitem{9196884}
D.~Barnes, M.~Gadd, P.~Murcutt, P.~Newman, and I.~Posner, ``The oxford radar
  robotcar dataset: A radar extension to the oxford robotcar dataset,'' in
  \emph{IEEE International Conference on Robotics and Automation (ICRA)}, 2020,
  pp. 6433--6438.

\bibitem{mur2015orb}
R.~Mur-Artal, J.~M.~M. Montiel, and J.~D. Tardos, ``Orb-slam: a versatile and
  accurate monocular slam system,'' \emph{IEEE transactions on robotics},
  vol.~31, no.~5, pp. 1147--1163, 2015.

\bibitem{liu2021balm}
Z.~Liu and F.~Zhang, ``Balm: Bundle adjustment for lidar mapping,'' \emph{IEEE
  Robotics and Automation Letters}, vol.~6, no.~2, pp. 3184--3191, 2021.

\bibitem{triggs2000bundle}
B.~Triggs, P.~F. McLauchlan, R.~I. Hartley, and A.~W. Fitzgibbon, ``Bundle
  adjustment—a modern synthesis,'' in \emph{Vision Algorithms: Theory and
  Practice: International Workshop on Vision Algorithms Corfu, Greece,
  September 21--22, 1999 Proceedings}.\hskip 1em plus 0.5em minus 0.4em\relax
  Springer, 2000, pp. 298--372.

\bibitem{besl1992method}
P.~J. Besl and N.~D. McKay, ``Method for registration of 3-d shapes,'' in
  \emph{Sensor fusion IV: control paradigms and data structures}, vol.
  1611.\hskip 1em plus 0.5em minus 0.4em\relax Spie, 1992, pp. 586--606.

\bibitem{dellenbach2022ct}
P.~Dellenbach, J.-E. Deschaud, B.~Jacquet, and F.~Goulette, ``Ct-icp: Real-time
  elastic lidar odometry with loop closure,'' in \emph{2022 International
  Conference on Robotics and Automation (ICRA)}.\hskip 1em plus 0.5em minus
  0.4em\relax IEEE, 2022, pp. 5580--5586.

\bibitem{li2020iterative}
J.~Li, C.~Zhang, Z.~Xu, H.~Zhou, and C.~Zhang, ``Iterative distance-aware
  similarity matrix convolution with mutual-supervised point elimination for
  efficient point cloud registration,'' in \emph{European conference on
  computer vision}.\hskip 1em plus 0.5em minus 0.4em\relax Springer, 2020, pp.
  378--394.

\bibitem{pan2021robust}
L.~Pan, Z.~Cai, and Z.~Liu, ``Robust partial-to-partial point cloud
  registration in a full range,'' \emph{arXiv preprint arXiv:2111.15606}, 2021.

\bibitem{moenning2004intrinsic}
C.~Moenning and N.~A. Dodgson, ``Intrinsic point cloud simplification,''
  \emph{Proc. 14th GrahiCon}, vol.~14, p.~23, 2004.

\bibitem{dovrat2019learning}
O.~Dovrat, I.~Lang, and S.~Avidan, ``Learning to sample,'' in \emph{Proceedings
  of the IEEE/CVF Conference on Computer Vision and Pattern Recognition}, 2019,
  pp. 2760--2769.

\bibitem{lang2020samplenet}
I.~Lang, A.~Manor, and S.~Avidan, ``Samplenet: Differentiable point cloud
  sampling,'' in \emph{Proceedings of the IEEE/CVF Conference on Computer
  Vision and Pattern Recognition}, 2020, pp. 7578--7588.

\bibitem{qi2017pointnet++}
C.~R. Qi, L.~Yi, H.~Su, and L.~J. Guibas, ``Pointnet++: Deep hierarchical
  feature learning on point sets in a metric space,'' \emph{Advances in neural
  information processing systems}, vol.~30, 2017.

\bibitem{chum2003locally}
O.~Chum, J.~Matas, and J.~Kittler, ``Locally optimized ransac,'' in
  \emph{Pattern Recognition: 25th DAGM Symposium, Magdeburg, Germany, September
  10-12, 2003. Proceedings 25}.\hskip 1em plus 0.5em minus 0.4em\relax
  Springer, 2003, pp. 236--243.

\bibitem{zeng20173dmatch}
A.~Zeng, S.~Song, M.~Nie{\ss}ner, M.~Fisher, J.~Xiao, and T.~Funkhouser,
  ``3dmatch: Learning local geometric descriptors from rgb-d reconstructions,''
  in \emph{Proceedings of the IEEE conference on computer vision and pattern
  recognition}, 2017, pp. 1802--1811.

\bibitem{pan2021variational}
L.~Pan, X.~Chen, Z.~Cai, J.~Zhang, H.~Zhao, S.~Yi, and Z.~Liu, ``Variational
  relational point completion network,'' in \emph{Proceedings of the IEEE/CVF
  conference on computer vision and pattern recognition}, 2021, pp. 8524--8533.

\bibitem{yuan2020deepgmr}
W.~Yuan, B.~Eckart, K.~Kim, V.~Jampani, D.~Fox, and J.~Kautz, ``Deepgmr:
  Learning latent gaussian mixture models for registration,'' in \emph{European
  conference on computer vision}.\hskip 1em plus 0.5em minus 0.4em\relax
  Springer, 2020, pp. 733--750.

\bibitem{yew2020rpm}
Z.~J. Yew and G.~H. Lee, ``Rpm-net: Robust point matching using learned
  features,'' in \emph{Proceedings of the IEEE/CVF conference on computer
  vision and pattern recognition}, 2020, pp. 11\,824--11\,833.

\bibitem{lin2022task}
Y.~Lin, L.~Chen, H.~Huang, C.~Ma, X.~Han, and S.~Cui, ``Task-aware sampling
  layer for point-wise analysis,'' \emph{IEEE Transactions on Visualization and
  Computer Graphics}, 2022.

\bibitem{liu2021deep}
S.-L. Liu, H.-X. Guo, H.~Pan, P.-S. Wang, X.~Tong, and Y.~Liu, ``Deep implicit
  moving least-squares functions for 3d reconstruction,'' in \emph{Proceedings
  of the IEEE/CVF Conference on Computer Vision and Pattern Recognition}, 2021,
  pp. 1788--1797.

\bibitem{segal2009generalized}
A.~Segal, D.~Haehnel, and S.~Thrun, ``Generalized-icp.'' in \emph{Robotics:
  science and systems}, vol.~2, no.~4.\hskip 1em plus 0.5em minus 0.4em\relax
  Seattle, WA, 2009, p. 435.

\end{thebibliography}

\end{document}